%% file: main.tex
\documentclass[10pt,twocolumn,letterpaper]{article}

\usepackage{cvpr}              

\input{preamble}

\definecolor{cvprblue}{rgb}{0.21,0.49,0.74}
\usepackage[pagebackref,breaklinks,colorlinks,allcolors=cvprblue]{hyperref}

\def\paperID{} 
\def\confName{}
\def\confYear{}

\usepackage[utf8]{inputenc} 
\usepackage[T1]{fontenc}    
\usepackage{hyperref}       
\usepackage{url}            
\usepackage{booktabs}       
\usepackage{amsfonts}       
\usepackage{nicefrac}       
\usepackage{microtype}      
\usepackage{xcolor}         

\usepackage{xspace}
\usepackage{dirtytalk}
\usepackage{graphicx} 
\usepackage{amsmath}
\usepackage{algorithm}
\usepackage{algpseudocode}
\usepackage{multirow}
\usepackage{array}
\usepackage{colortbl} 
\usepackage{xcolor}   
\usepackage{soul}
\usepackage{caption}

\DeclareRobustCommand{\ctext}[2]{{\sethlcolor{#1}\hl{#2}}}

\definecolor{highcolor}{HTML}{D7EDEB}
\definecolor{midcolor}{HTML}{FAF0DA}
\definecolor{lowcolor}{HTML}{FADDD6}
\definecolor{conditioncolor}{HTML}{FFDD7D}
\definecolor{statecolor}{HTML}{FAAC6C}
\definecolor{activitycolor}{HTML}{93ABD9}
\definecolor{othercolor}{HTML}{A7B0C2}

\newcommand{\condition}[1]{\ctext{conditioncolor}{#1}}
\newcommand{\state}[1]{\ctext{statecolor}{#1}}
\newcommand{\activity}[1]{\ctext{activitycolor}{#1}}
\newcommand{\other}[1]{\ctext{othercolor}{#1}}

\newcommand{\high}[1]{\ctext{highcolor}{#1}}
\newcommand{\moderate}[1]{\ctext{midcolor}{#1}}
\newcommand{\low}[1]{\ctext{lowcolor}{#1}}

\definecolor{lightgray}{gray}{0.8}

\newcommand{\ours}{ELSA\xspace}
\input{math_commands.tex}

\title{ELSA: Evaluating Localization of Social Activities \\ in Urban Streets using Open-Vocabulary Detection}

\author{%
\textbf{Maryam Hosseini}$^{1*}$ \quad \textbf{Marco Cipriano}$^{2*}$ \quad  \textbf{Daniel Hodczak}$^{3*}$ \quad \textbf{Sedigheh Eslami}$^2$ \\ \textbf{Liu Liu$^1$} \quad\textbf{Andres Sevtsuk}$^1$ \quad \textbf{Gerard de Melo}$^2$\\
$^1$Massachusetts Institute of Technology (MIT) \\ $^2$ Hasso Plattner Institute (HPI) \\ $^3$University of Illinios Chicago (UIC)\\
\texttt{maryamh@mit.edu, marco.cipriano@hpi.de}\\
}

\begin{document}
\twocolumn[{%
\renewcommand\twocolumn[1][]{#1}%
\maketitle

\includegraphics[width=\linewidth]{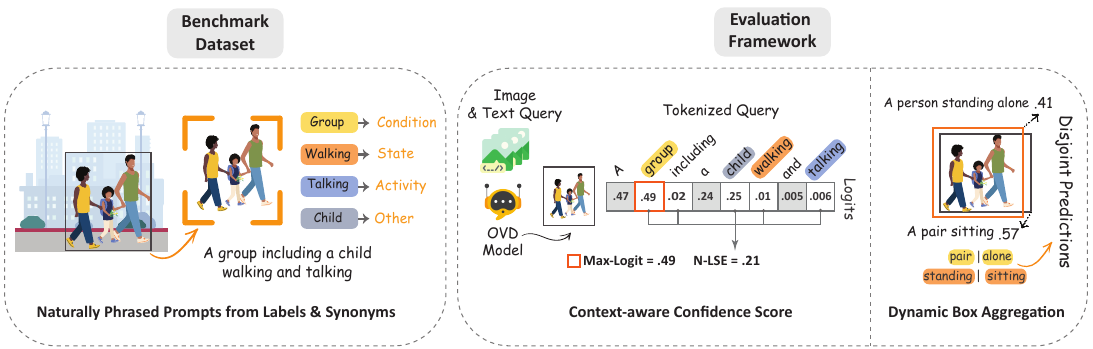}
\captionof{figure}{We present \ours: Evaluating Localization of Social Activities—a novel \emph{benchmark dataset} and \emph{evaluation framework} for assessing open-vocabulary detection (OVD) models in recognizing and localizing social interactions on urban streets from still images. \ours includes a multi-label annotation scheme spanning four categories:~\condition{Condition},~\state{State},~\activity{Activity}, and~\other{Other}. It features natural language prompts derived from these labels, along with synonymous variations to rigorously test models' semantic comprehension. Our \emph{N-LSE} context-aware confidence score surpasses max-logit scoring, yielding more realistic confidence scores and effectively reducing false positives. Our \emph{DBA} algorithm dynamically groups overlapping predictions, ensuring semantic coherence and recovering correct predictions that otherwise would be missed by class-agnostic NMS.
\vspace{1em}}
\label{fig:teaser}
}]

\makeatletter\def\Hy@Warning#1{}\makeatother
\def\thefootnote{*}\footnotetext{Equal contribution.}
\def\thefootnote{\arabic{footnote}}

\input{sec/00-abstract}
\input{sec/01-intro}
\input{sec/02-related}
\input{sec/03-method}
\input{sec/04-results}

\input{sec/06-conclusion}

 \small \bibliographystyle{ieeenat_fullname} \bibliography{main}

\clearpage
\input{sec/07-appendix}

\end{document}

%% file: preamble.tex
\newcommand{\algname}{DBA\xspace} 

%% file: math_commands.tex
\usepackage{amsmath,amsfonts,bm}

\def\eqref#1{equation~\ref{#1}}

\def\1{\bm{1}}

\DeclareMathAlphabet{\mathsfit}{\encodingdefault}{\sfdefault}{m}{sl}
\SetMathAlphabet{\mathsfit}{bold}{\encodingdefault}{\sfdefault}{bx}{n}



%% file: sec/00-abstract.tex
\begin{abstract}
Existing Open Vocabulary Detection (OVD) models exhibit a number of challenges. They often struggle with semantic consistency across diverse inputs, and are often sensitive to slight variations in input phrasing, leading to inconsistent performance.  The calibration of their predictive confidence, especially in complex multi-label scenarios, remains suboptimal, frequently resulting in overconfident predictions that do not accurately reflect their context understanding. To understand these limitations, multi-label detection benchmarks are needed. A particularly challenging domain for such benchmarking is social activities. Due to the lack of multi-label benchmarks for social interactions, in this work we present \ours: Evaluating Localization of Social Activities. \ours draws on theoretical frameworks in urban sociology and design and uses in-the-wild street-level imagery, where the size of groups and the types of activities vary significantly. ELSA includes more than 900 manually annotated images with more than 4,300 multi-labeled bounding boxes for individual and group activities. We introduce a novel confidence score computation method NLSE and a novel Dynamic Box Aggregation (DBA) algorithm to assess semantic consistency in overlapping predictions. We report our results on the widely-used SOTA models Grounding DINO, Detic, OWL, and MDETR. Our evaluation protocol considers semantic stability and localization accuracy and further exposes the limitations of existing approaches. 
\end{abstract}

%% file: sec/01-intro.tex
\section{Introduction}
\label{sec:intro}
Recently, increased focus on the human scale of the cities has drawn more attention to public spaces and pedestrian facilities.For decades, urban scholars have been fascinated by the complex interplay between public spaces and the social interactions they support~\cite{mehta2021revisiting, whyte1980social, jacobs1961death}. However, scientific inquiry into the distribution of social activities has been hampered by data collection costs and time requirements.

The emergence of advanced computer vision techniques and the availability of public sources of street-level imagery have opened new avenues for conducting comprehensive observational studies at reduced cost and increased scale. Activity recognition techniques are mostly designed to work with videos~\cite{lan2012social}, since, by nature, human activity involves motion and sequence of actions. Yet, acquiring continuous video footage across an entire city over time entails arduous data storage requirements and processing costs. Object detection on still images emerges as a low-cost, efficient, and applicable method, as it allows for the identification and localization of complex social interactions in diverse settings, where the environmental context significantly influences the range of possible social interactions and each image can contain a large number of people engaged in diverse activities.

While conventional object detection models are trained in closed-vocabulary settings and rely heavily on predefined classes, open-vocabulary detection (OVD) models aim to transcend traditional object detection models, and utilize the abundance of language data to facilitate the detection of uncommon classes in standard benchmark training data. A robust OVD model is expected to handle a wide range of input terms and phrases that were not explicitly part of its training set. This is crucial for models deployed in real-world settings, such as urban streets, where unpredictable and varied interactions are common. The absence of benchmark data for OVD of social and individual actions in still images `in the wild' hinders the development of models that generalize well across diverse and spontaneous urban scenarios, where the context and variability of human activities are far greater than those typically encountered in controlled environments. Furthermore, OVDs pose new challenges in both localization and semantic understanding of unseen new categories. They often struggle with semantic consistency across diverse inputs, are sensitive to slight variations in input phrasing, and have suboptimal predictive confidence calibration in out-of-distribution scenarios, resulting in overconfident predictions that do not accurately reflect their actual accuracy~\cite{schulter2023omnilabel,dave2021evaluating}.

In response to these challenges, we propose \ours, a new benchmark dataset and evaluation framework to evaluate the performance of OVD models in recognizing and localizing human activity in urban streets from still images. Our dataset employs a multi-label annotation scheme encompassing 33 distinct labels that can be concurrently assigned to each bounding box. This results in over 4,000 bounding boxes annotated with 115 unique combinations of human activities across 934 street view images. To enhance the evaluation process, we have generated precise, naturally phrased sentences for each label combination and their near synonyms, totaling 830 unique prompts. These prompts were applied to each image during evaluation, providing a comprehensive and nuanced assessment of the models' ability to handle varied and contextually rich descriptions of human activities.

Recognizing the intertwined nature of OVD models with language features—and the evaluation complexities this presents—we introduce Dynamic Box Aggregation (DBA), a method designed to address overlapping detections and disjoint predictions in open-vocabulary models. Unlike Non-Maximum Suppression (NMS), DBA retains predictions with confidence scores near the maximum within a specified threshold, while identifying and penalizing disjoint predictions for the same bounding box. DBA integrates seamlessly into evaluation metrics like Average Precision (AP) and F1-score, enhancing both localization and semantic understanding. This provides a more nuanced approach compared to traditional NMS techniques and their variants, such as NMS-AP~\cite{yao2024evaluate}. Additionally, we propose a novel method to compute the confidence score, N-LSE, for phrase grounding and referral expression. N-LSE prioritizes predictions based on the most salient sub-phrases and tokens within the query, effectively aligning token-level language features with visual features in the predicted regions.

Our contributions are threefold: \textbf{1)} We present \ours, the first benchmark dataset tailored for evaluating OVD models in the context of human activity recognition in urban street images with an extensive set of 830 naturally phrased prompts encompassing various label combinations and synonyms, enhancing the robustness and depth of the evaluation. \textbf{2)} We expand the evaluation focus beyond traditional metrics by identifying and addressing three key areas that OVD models should target in their development and evaluation protocols: semantic understanding, semantic stability, and precise localization. \textbf{3)} We introduce the context-aware confidence score N-LSE to improve the alignment between the language queries and visual detection. Furthermore, we address the inflated AP problem, raised in previous works, by proposing a novel algorithm (DBA), enhancing both spatial localization and semantic accuracy. This method uniquely penalizes disjoint predictions on the same box, offering an approach that captures subtle nuances in object detection and surpasses the limitations of NMS-AP. 

These innovations address key challenges in OVD model evaluation and offer valuable tools for advancing future research in this domain.

%% file: sec/02-related.tex
\section{Related Work}
\label{sec:related}
\label{sec:planning}
\textbf{Social Interactions in Public Spaces.} Vibrant streets rich in interpersonal exchange have fascinated urban scholars because of their social qualities as well as fundamental indicators of sustainable urban environments~\cite{mehta2021revisiting}. William Whyte~\cite{whyte1980social} along with~\citet{jacobs1961death} highlight the intrinsic value of public spaces in fostering vibrant social life. 
Jan~\citet{gehl1968people} describes activities in the public spaces as a spectrum between optional activities, such as talking with friends, and necessary activities, like walking to work.  
The public space observational method~\cite{gehl2013study} delineates between active social group activities, like dining or talking together, and passive activities, such as strangers sitting on a bench checking their cell-phones.  
Inspired by this research, we define the target set of social activities in \ours. 

\textbf{Open-Vocabulary Object Detection.} OVD, first introduced by \citet{zareian2021open}, primarily tackles the limitation of traditional object detection models that rely on pre-defined closed set of objects \cite{bravo2023open, minderer2024scaling, kim2023detection} tested on various OVD benchmark datasets \cite{schulter2023omnilabel, yao2024evaluate}. At their core, a vision-language contrastive loss is often used for aligning semantics and concepts in the two modalities \cite{kamath2021mdetr, liu2023grounding, cheng2024yolo, h-detr, owlvit, li2022grounded} with additional soft token prediction in MDETR \cite{kamath2021mdetr}. Using a dual-encoder-single-decoder architecture, Grounding DINO~\cite{liu2023grounding} extends DINO \cite{dino} such that given a text prompt, query selection is performed to select the text features relevant to the cross-modal decoder. A contrastive loss for aligning the output of the decoder and text queries along with a regression L1 loss and generalized union over intersection is optimized end-to-end for the detection. Detic~\cite{zhou2022detecting} trains the classifier of a image detector on classification data, expanding the detector's vocabulary to tens of thousands of concepts. In OWL-ViT~\cite{owlvit}, the fixed classification weights of a pre-trained Vision Transformer are replaced with text embeddings generated by the pre-trained encoder.

\textbf{OVD Evaluation.} 
The standard evaluation metric for object detection is the mean of the per-class average precision (mAP)~\cite{everingham2010pascal}. 
As shown by~\citet{dave2021evaluating}, standard AP is sensitive to changes in cross-category ranking. Furthermore, \citet{yao2024evaluate} show the inflated AP problem and proposes to suppress it using class-ignored NMS-AP that unifies multiple predictions of the same box and assigns the most confident label to that box. 
Relying on the maximum-logit confidence, this method is also prone to misrepresenting the correct ranking of relevant boxes and can inaccurately represent the robustness and stability of the model in predicting the correct class, as it is merely relies on the maximum-logit token from the query.
In contrast, our approach ranks the predicted boxes with respect to all tokens in the query, which is crucial for multi-label scenarios.

\textbf{Activity Localization Datasets.} Activity localization 
involves analyzing the activities in a sequence of images~\cite{Bagautdinov_2017_CVPR, barekatain2017okutama, ehsanpour2022jrdb, zhou_human_2022, zhou_human_2022}.  A seminal study by \citet{choi2009they} focuses on in-the-wild pedestrian action classification from videos. 
 Recent advancements in~\citet{zhou_human_2022} and~\citet{wang_human--human_2023}
 combine appearance and pose data with transformers to enhance interaction recognition and improve the detection of complex human behaviors.
 \citet{li_hake_2023} introduced cognitive depth with the HAKE engine, leveraging logical reasoning to analyze human--object interactions. Most existing models are tested on video datasets, including Volleyball~\cite{ibrahim2016hierarchical}, AVA-Interaction~\cite{wang_human--human_2023}, HICO-DET~\cite{chao2018learning}, V-COCO~\cite{v-coco}, NTU RGB+D~\cite{shahroudy2016ntu, liu2019ntu}, SBU-Kinect-Interaction~\cite{yun2012two}, and MatchNMingle~\cite{cabrera2018matchnmingle}. \citet{ehsanpour2022jrdb} introduced JRDB-Act, a video dataset for group-based social activities in university campus scenes. In contrast, \ours focuses on the more challenging task of localizing social activities in images, where models must infer activities from a snapshot without the temporal cues available in videos.









%% file: sec/03-method.tex
\section{\ours: A Benchmark for Evaluating Localization of Social Activities}
\label{sec:method}
Despite advances in object detection, existing benchmarks inadequately address the detection of nuanced human activities and social interactions in dynamic urban environments. \ours introduces a comprehensive dataset designed to bridge this gap by providing annotated instances of diverse activities, group behaviors, and challenging urban scenarios traditional object detection models struggle with.

\textbf{Image Resources.} We selected New York City as our site of interest due to its vibrant streets and public spaces. We compiled street-level images from two different sources: Microsoft Bing Streetside~\cite{kopf2010street} and Google Street View~\cite{gsvapi, anguelov2010google}. The Bing imagery provides time-stamps, making it possible to choose days and times with a higher probability of encountering pedestrians on the streets.

\begin{figure*}[t]
\begin{center}
\includegraphics[width=0.8\linewidth]{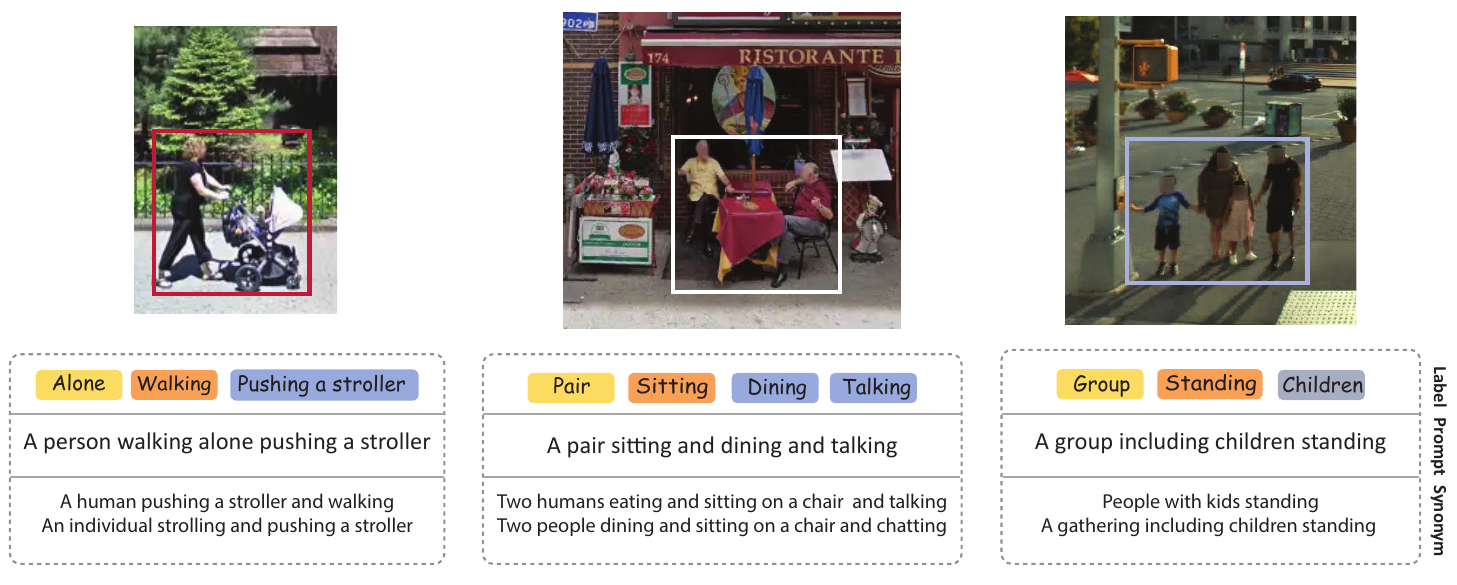}
\end{center}
   \caption{Examples of individual annotations extracted from larger images in the benchmark. Each bounding box is accompanied by a set of labels, a base natural language prompt, and a series of synonyms, two of which are shown here.}
\label{fig:annotations}
\end{figure*}

\textbf{Target Labels } 
We draw on the literature on active design and urban vibrancy (see Section~\ref{sec:planning}) to select our primary individual labels. 
\ours exhibits non-disjoint label spaces, where multiple concurrent labels can be applied to the same object in a multi-labeling scheme that encompasses the nuances of human behavior and context. Labels are grouped into three main categories: 1)  \condition{Condition:} defines the social configuration of the subjects as \emph{alone}, \emph{two people}, or \emph{group}. These labels are disjoint and denote mutually exclusive social settings, establishing the primary context for potential interactions, such as solo activities, limited interactions, or group dynamics; 2) \state{State:} captures the physical disposition or activity mode of the subjects, such as \emph{walking} or \emph{sitting}. While disjoint for individuals, these labels can co-occur in couple or group scenarios, indicating stationary engagement (\emph{standing}, \emph{sitting}) or transient interactions (\emph{walking}, \emph{biking}); 3) \activity{Activity:} reflects specific behaviors or activities, such as \emph{dining} or \emph{talking}; 4) \other{Other:} any other information that can be of interest in urban activity analysis such as the presence of children, dogs, luggage, coffee or drink, etc. We report additional information about the label categories in Appendix~\ref{subsec:label-categories}.

\textbf{Annotation Process.} We customized the open source Label Studio tool \cite{label_studio} for annotation and integrated YOLOv8 \cite{yolov8_ultralytics} for pre-detecting the initial objects. A team of four trained annotators manually corrected the initial boxes and annotated the label combinations according to predefined guidelines. An urban planning expert then reviewed all annotations to ensure accuracy and consistency.

Examples of \ours's annotations are depicted in Figure~\ref{fig:annotations}. Additional examples are included in Appendix~\ref{subsec:additional-annotations}.

\textbf{Annotation Cleaning.} Following the initial annotations, we applied sanity checks to ensure label consistency. For example, a single person cannot be labeled as both \emph{sitting} and \emph{walking}. The complete list of sanity rules is provided in Appendix~\ref{subsec:full-label-list}. Annotations failing these checks were re-evaluated and corrected, and this process was repeated until all boxes met the defined criteria.

\begin{figure}[hb!]
    \centering
    \includegraphics[width=0.9\linewidth]{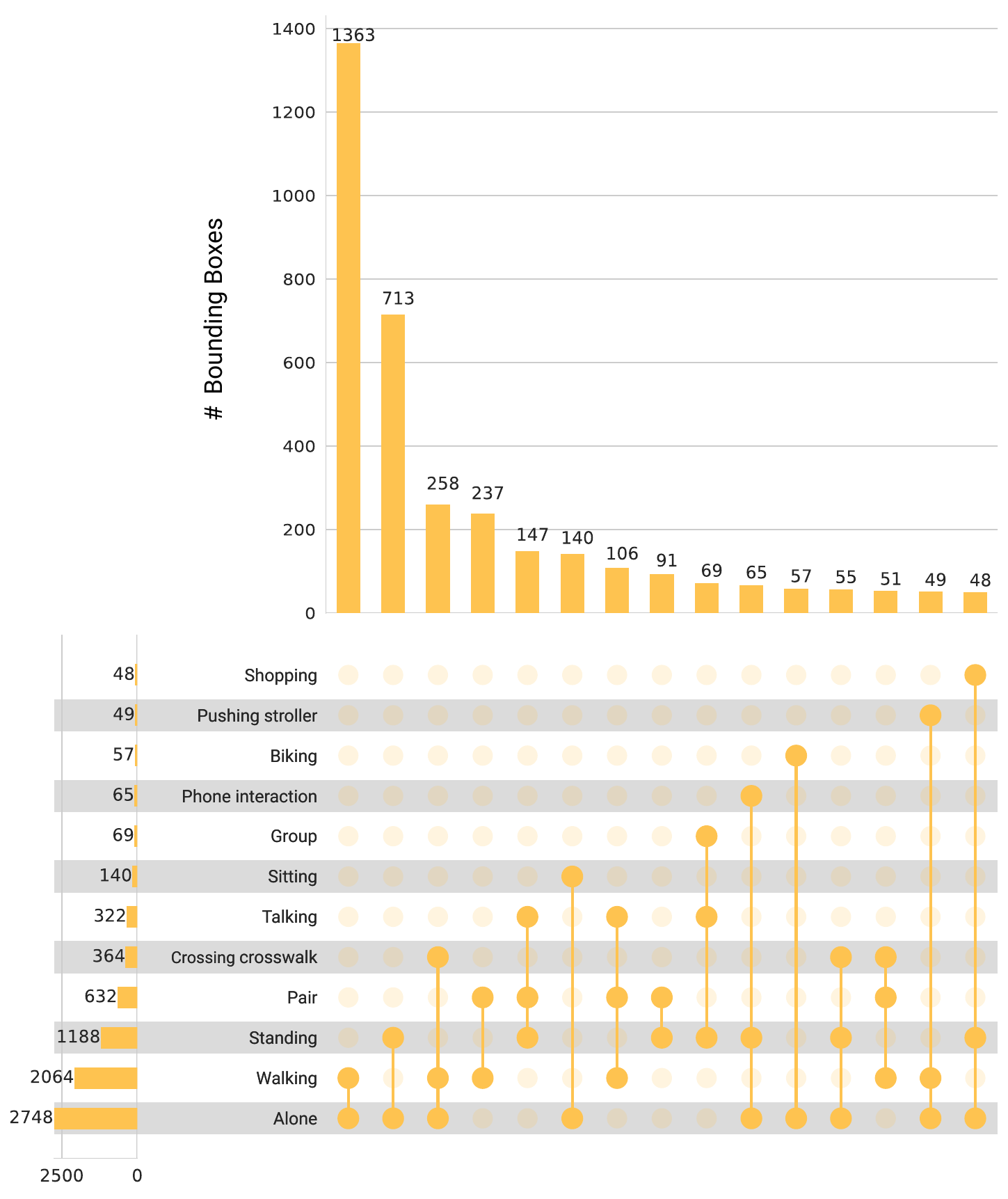}
    \caption{Overview of label distribution and combinations in \ours, showing the 15 most frequently occurring label combinations. Connected dots represent label combinations, with frequencies plotted in the bar charts above each combination. For example, the ``walking alone'' combination appears 1,363 times, while ``standing alone'' appears 713 times.}
    \label{fig:data-distributions}
\end{figure}

Dataset Statistics. \ours consists of 934 images with over 4.3K annotated bounding boxes for social and individual activities, encompassing 34 distinct labels. Each box averages 2.5 labels, with `walking' and `alone' being the most frequent. This results in 114 unique combinations of human activities in the dataset. Figure~\ref{fig:data-distributions} displays the distribution of individual labels and their combinations in \ours.

\textbf{Prompt Formation.} Unlike physical objects, activities and human--human or human--object interactions pose significant challenges in being accurately captured by a single word or label. To investigate this, we conducted a series of tests on various models, examining their responses to prompts with verbs like “walking,” “talking,” or “standing,” and phrases like “walking alone” or “talking in groups.” As expected, the results were often inaccurate or non-existent. These models require more detailed natural language descriptions to detect these activities correctly, such as “an individual sitting on a bench.”
To address this need, we augmented \ours by generating precise, naturally phrased sentences for each label combination and their near synonyms. This capability ensures that the models receive comprehensive descriptions, significantly improving their detection accuracy.

\subsection{Evaluation Framework}

Evaluating open-vocabulary detection (OVD) models on complex human activities and interactions presents unique challenges. Unlike traditional object detection tasks, OVD models must interpret rich natural language queries and accurately localize corresponding instances in images. In this section, we introduce a robust evaluation methodology designed to assess both the semantic understanding and localization accuracy of OVD models in multi-label scenarios 
An effective OVD model should excel in these two key aspects: 1) \textbf{Semantic Understanding}: Accurately interpret detailed query phrases to detect the correct targets, 
2) \textbf{Localization Accuracy}: Precisely localize target objects or interactions in images by effectively bridging natural language and visual features.

\subsubsection{Context-aware Confidence Score}
Unlike single-object detection, multi-label human activity and interaction detection presents additional challenges for identifying multiple overlapping targets, activities, and interactions within the same scene. Thus, boxes must reflect not only the presence of the targets but also their attributes such as state and condition with higher confidence.
In open-vocabulary detection, specifically, phrase grounding models, each predicted box is typically associated with a confidence score and an array of logits. These logits quantify the model's confidence in the relationship between the visual features within the box and specific tokens. Often, the confidence score of a box is determined by the highest logit value, i.e., Max-Logit, among all tokens \cite{liu2023grounding}. While effective in some contexts, the Max-Logit approach can bias the confidence towards prevalent object classes (e.g., ``person''), potentially overlooking nuanced attributes critical in multi-label scenarios.

To address the limitations inherent in these methods—such as biases towards prevalent classes or insufficient emphasis on nuanced attributes—we introduce a \emph{Context-Aware Confidence Score} that aggregates the logits of selected relevant tokens to compute a more representative confidence score. Specifically, we propose the \textbf{Normalized Log-Sum-Exp (N-LSE)} function over tokens as:

\begin{equation}
   \text{N-LSE}(\mathbf{z}) = \log \left(\frac{1}{N} \sum_{t=1}^{N} e^{z_t} \right) = \log \left( \sum_{t=1}^{N} e^{z_t} \right) - \log(N),
\label{eq:lse}
\end{equation}

Here, \(\mathbf{z}\) represents the vector of logits, and \(N\) is the number of elements (corresponding to each token) in \(\mathbf{z}\).

Our scoring method involves mapping the tokens to boxes and comparing the native scoring of each model like Max-Logit in Grounding DINO with N-LSE methods. By capturing the combined evidence from pertinent tokens, our approach mitigates the bias towards any single prevalent token~\cite{liu2023grounding} or the limitations of a ``no object'' probability~\cite{kamath2021mdetr}. This not only decreases false positives but also retrieves under-ranked boxes.

Following re-ranking, we apply a confidence threshold to prune low-confidence predictions. Based on prior work~\cite{scalingmindrer} and empirical validation, we set the N-LSE confidence threshold to 0.3. Predicted bounding boxes with N-LSE scores below this threshold are discarded.
To manage overlapping predictions, for each ground truth (GT) bounding box, we identify the predicted bounding box with the highest Intersection over Union (IoU) relative to that GT box—designated as the \emph{anchor}. Other predicted boxes with an IoU greater than 0.85 relative to the anchor are grouped together, forming a cluster of candidates representing the same object or interaction. This grouping allows us to consider multiple high-confidence predictions that may correspond to the same GT instance.

\subsubsection{Dynamic Box Aggregation (DBA)}

\begin{figure}
    \centering
    \includegraphics[width=\linewidth]{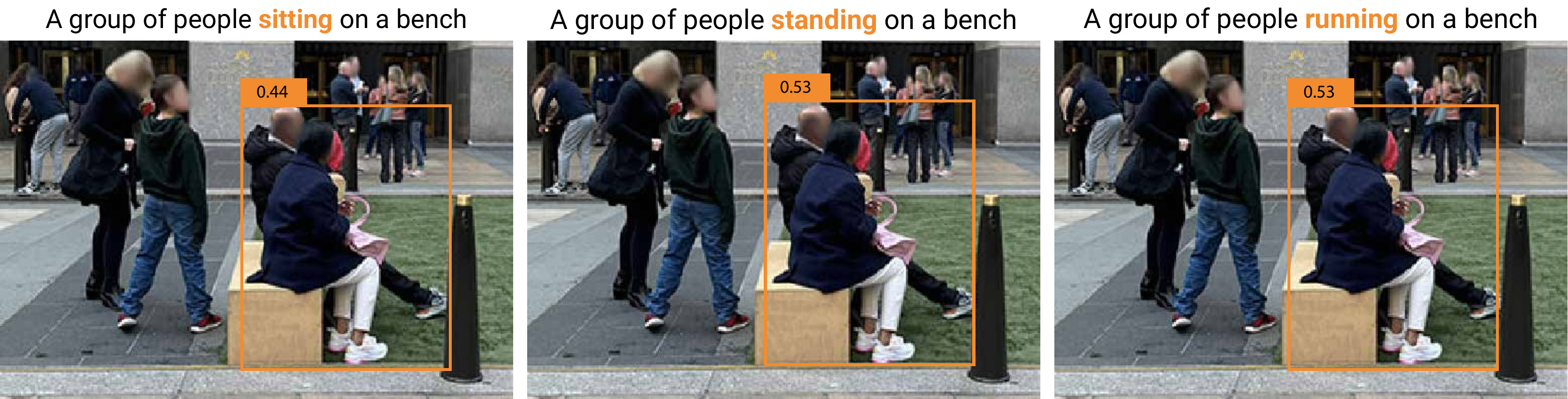}
    \caption{Using the Grounding DINO model with Swin-T backbone and Max-logit scoring to run variations of the same prompt with different \state{states}. }
    \label{fig:qual}
\end{figure}

A significant challenge in evaluating OVD models is handling multiple predictions for the same object arising from different prompts or overlapping boxes. \citet{yao2024evaluate} proposed combining class-ignored non-maximum suppression (C-NMS) with average precision computation, which, for any set of overlapping prediction boxes, selects the box with the highest confidence, suppresses the rest, and computes AP over the subset of the prediction boxes instead.

While more effective than the traditional AP, this approach has notable drawbacks: 1) It does not expose the model's susceptibility to making disjoint predictions with close confidence levels, and 2) It may incorrectly suppress true positives with confidence levels close to the highest prediction as false positives, as also raised by the original paper~\cite{yao2024evaluate}. 

To address the limitations of existing methods like NMS-AP, we introduce the \emph{Dynamic Box Aggregation} (DBA) (Algorithm~\ref{alg:dba}), which considers both confidence scores and semantic coherence by considering the disjoint predictions for the same object among overlapping boxes. Unlike methods that rely solely on C-NMS, DBA retains boxes with confidence scores within a specified threshold (\texttt{score\_thr}) of the maximum, suppressing the rest. To ensure optimal performance, the threshold for confidence retention in our DBA algorithm is dynamically determined for each model. Specifically, we perform an extensive hyperparameter sweep across a range of thresholds and select the value that maximizes the precision-recall balance, thereby adapting DBA to the confidence calibration of each model.

An important component of DBA is its ability to identify disjoint predictions with similar confidence scores. It exposes the model's vulnerability in understanding the target by penalizing cases where multiple inconsistent predictions are made for the same bounding box. For example, DBA would flag and penalize disjoint predictions like a person being detected as ``sitting," ``walking," and ``running" with close confidence such as the one shown in Figure~\ref{fig:qual}. This approach enhances evaluation accuracy by highlighting such failures in semantic understanding, which are otherwise overlooked by traditional C-NMS methods, ensuring that only predictions with sufficiently high confidence are considered, reducing false positives while maintaining high recall.

\begin{algorithm}[h!]
\caption{Dynamic Box Aggregation (DBA)}
\label{alg:dba}
\begin{algorithmic}[1]
\State \textbf{Input:} Grouped bboxes $O$, ground truth set GT, IoU threshold $\text{iou\_thr}$, score threshold $\text{score\_thr}$
\State \textbf{Output:} TP, FP

\State Initialize empty dictionary data structures for TP, FP

\For{each group $o \in O$}
    \State $T = \max(\text{Scores}(o)) - \text{score\_thr}$
    \For{each bbox $b_i$ in grp $o$}
        \If{$\text{Score}(b_i) \geq T$}
            \If{predicted labels are disjoint in condition or state}
                \State Add $b_i$ to the FP 
            \Else
                \If{$\text{IoU}(b_i, g) \geq \text{iou\_thr}$ and $\text{label}(b_i) \subseteq \text{label}(g)$ for any $g \in GT$}
                    \State Add $b_i$ to the TP 
                \Else
                    \State Add $b_i$ to the FP 
                \EndIf 
            \EndIf
        \EndIf
    \EndFor
\EndFor

\State \Return TP, FP 

\end{algorithmic}
\end{algorithm}

\algname handles overlapping predictions, while non-overlapping predictions are processed separately and later combined with DBA results for final metric computation, as shown in Algorithm~\ref{alg:cdba_apf}. 

\begin{algorithm}[h!]
\caption{\algname Integrated Evaluation}
\label{alg:cdba_apf}
\begin{algorithmic}[1]
\State \textbf{Input:} Non-overlapping bounding boxes $N$, GT, TP and FP from \algname, IoU threshold $\text{iou\_thr}$
\State \textbf{Output:} AP and F$_1$ score
\State Initialize empty dictionary data structure for FN
\State Initialize empty dictionary data structure for matched GT $matchedGT$

\For{each $n \in N$}
    \If{$\text{IoU}(n, g) \geq \text{iou\_thr}$ and $\text{label}(n) \subseteq \text{label}(g)$ for any $g \in GT$}
        \State Add $n$ to TP, update $matchedGT$
    \Else
        \State Add $n$ to FP
    \EndIf
\EndFor

\For{each $g \in GT$ not in $matchedGT$}
    \State Add $g$ to FN
\EndFor

\State Compute precision and recall at each threshold
\State Compute AP 

\State \Return AP, F$_1$

\end{algorithmic}
\end{algorithm}

In summary, \ours provides a challenging and comprehensive benchmark for evaluating the detection of social interactions and individual activities in complex urban scenes. By addressing both visual data and prompt-level challenges, and introducing novel evaluation methods like context-aware re-ranking and \algname, our benchmark facilitates a deeper understanding of model performance in open-vocabulary detection tasks. In the following section, we present experimental results demonstrating the effectiveness of our approach.

%% file: sec/04-results.tex
\begin{table*}[h!]
\caption{Comparison of baseline mAP (without C-NMS or DBA) using our proposed scoring function N-LSE and  maximum confidence score among all tokens, at four levels: global, CS (Condition + State), CSA (Condition + State + Activity), and CSO (Condition + State + Other). All scores are in percentage.}
\label{tab:resranking}
\begin{center}
\arrayrulecolor{lightgray} 
\begin{tabular}{l l l l l l l}
\rowcolor{lightgray!20} 
\textbf{Model} & \textbf{Variation} & \textbf{Ranking} & \textbf{Global} & \textbf{CS} & \textbf{CSA} & \textbf{CSO} \\ \arrayrulecolor{lightgray} \hline

\multirow{6}{*}{Gdino} 
          & Swin-B      & N-LSE  & \moderate{2.32E-01} & \moderate{6.76E-01} & \moderate{1.89E-02} & \moderate{3.15E-03} \\ \cline{2-7}
          &             & Max   & 1.99E-01 & 5.16E-01 & 9.93E-03 & 1.73E-03 \\ \cline{2-7}
          & Swin-T (1)  & N-LSE  & \moderate{9.69E-02} & \moderate{1.90E-01} & \moderate{3.25E-03} & \moderate{7.83E-04} \\ \cline{2-7}
          &             & Max   & 9.63E-02 & 1.81E-01 & 3.22E-03 & 6.39E-04 \\ \cline{2-7}
          & Swin-T (2)  & N-LSE  & \moderate{1.00E-01} & \moderate{2.56E-01} & 1.13E-03 & \moderate{4.30E-03} \\ \cline{2-7}
          &             & Max   & 9.58E-02 & 2.23E-01 & \moderate{3.56E-03} & 5.00E-04 \\ \hline

\end{tabular}
\end{center}
\end{table*}

\begin{table*}[t!]
\caption{Performance comparison of NMS-AP and DBA-AP using N-LSE confidence metrics at four levels: global, CS (Condition + State), CSA (Condition + State + Activity), and CSO (Condition + State + Other). The two variants of Swint-T are without $(1)$ and with $(2)$ finetuning on COCO. All scores are in percentage.}
\label{tab:resdba}
\begin{center}
\arrayrulecolor{lightgray} 
\begin{tabular}{l l l l l l l}
\rowcolor{lightgray!20} 
\textbf{Model} & \textbf{Variation} & \textbf{Scoring} & \textbf{Global} & \textbf{CS} & \textbf{CSA} & \textbf{CSO} \\ \arrayrulecolor{lightgray} \hline

\multirow{6}{*}{Gdino} 
                      & Swin-B      & DBA-AP  & \low{1.67E-01} & \low{4.06E-01} & \high{8.63E-03} & \low{3.00E-03} \\ \cline{2-7}
                      &             & NMS-AP  & 5.71E-01 & 1.05E+00 & 0 & 2.24E-02 \\ \cline{2-7}
                      & Swin-T (1)  & DBA-AP  & \low{7.82E-03} & \low{1.42E-02} & \low{5.87E-04} & \high{1.91E-03} \\ \cline{2-7}
                      &             & NMS-AP  & 3.73E-01 & 1.04E+00 & 3.17E-02 & 0 \\ \cline{2-7}
                      & Swin-T (2)  & DBA-AP  & \low{5.32E-02} & \high{1.23E-01} & \high{3.20E-04} & \high{1.90E-03} \\ \cline{2-7}
                      &             & NMS-AP  & 2.26E-01 & 0 & 0 & 0 \\ \hline
\multirow{4}{*}{MDETR} 
                      & EfficientNet & DBA-AP  & \low{3.76E-04} & \high{9.12E-04} & \low{4.15E-06} & \high{5.39E-07} \\ \cline{2-7}
                      &              & NMS-AP  & 6.18E-04 & 0 & 1.14E-05 & 0 \\ \cline{2-7}
                      & ResNet 101   & DBA-AP  & \low{3.04E-04} & \high{7.09E-04} & \high{9.96E-06} & \high{1.82E-07} \\ \cline{2-7}
                      &              & NMS-AP  & 2.60E-02 & 0 & 0 & 0 \\ \hline
\multirow{2}{*}{OWL} 
                      & ViT 32     & DBA-AP  & \low{1.25E-01} & \high{4.77E-01} & \high{6.35E-02} & \high{5.08E-06} \\ \cline{3-7}
                      &              & NMS-AP  & 2.06E-01 & 4.51E-01 & 0 & 0 \\ \cline{2-7} \hline
\multirow{2}{*}{Detic} 
                      & ResNet 50     & DBA-AP  & \high{4.53E-02} & \high{2.39E-01} & \high{6.92E-03} & \high{1.13E-03} \\ \cline{3-7}
                      &              & NMS-AP  & 2.13E-02 & 0 & 0 & 0 \\ \cline{2-7} \hline
\end{tabular}
\end{center}
\end{table*}

\section{Results}
\label{sec:res}

In this section, we compare our re-ranking strategy using N-LSE and the Max-Logit method used in prior work \cite{liu2023grounding, cheng2024yolo, li2022grounded}, discussing how N-LSE addresses key issues related to confidence mappings between logits and bounding boxes.

Next, we highlight the benefits provided by our DBA method across all benchmark levels, using the four state-of-the-art models Grounding DINO, MDETR, Detic, and OWL, across different backbones, for a total of seven different variants.

Finally, we discuss the general zero-shot performance of each model on our benchmark.

\subsection{Context-aware Confidence Score}

Grounding DINO has a limit of 900 predictions per image. For our dataset, comprising 934 images, we retrieved all 900 bounding boxes per image and applied a total of 830 prompts to each image. This process yields a total of 697,698,000 bounding boxes. 

\begin{figure}
\begin{center}
\includegraphics[width=\linewidth]{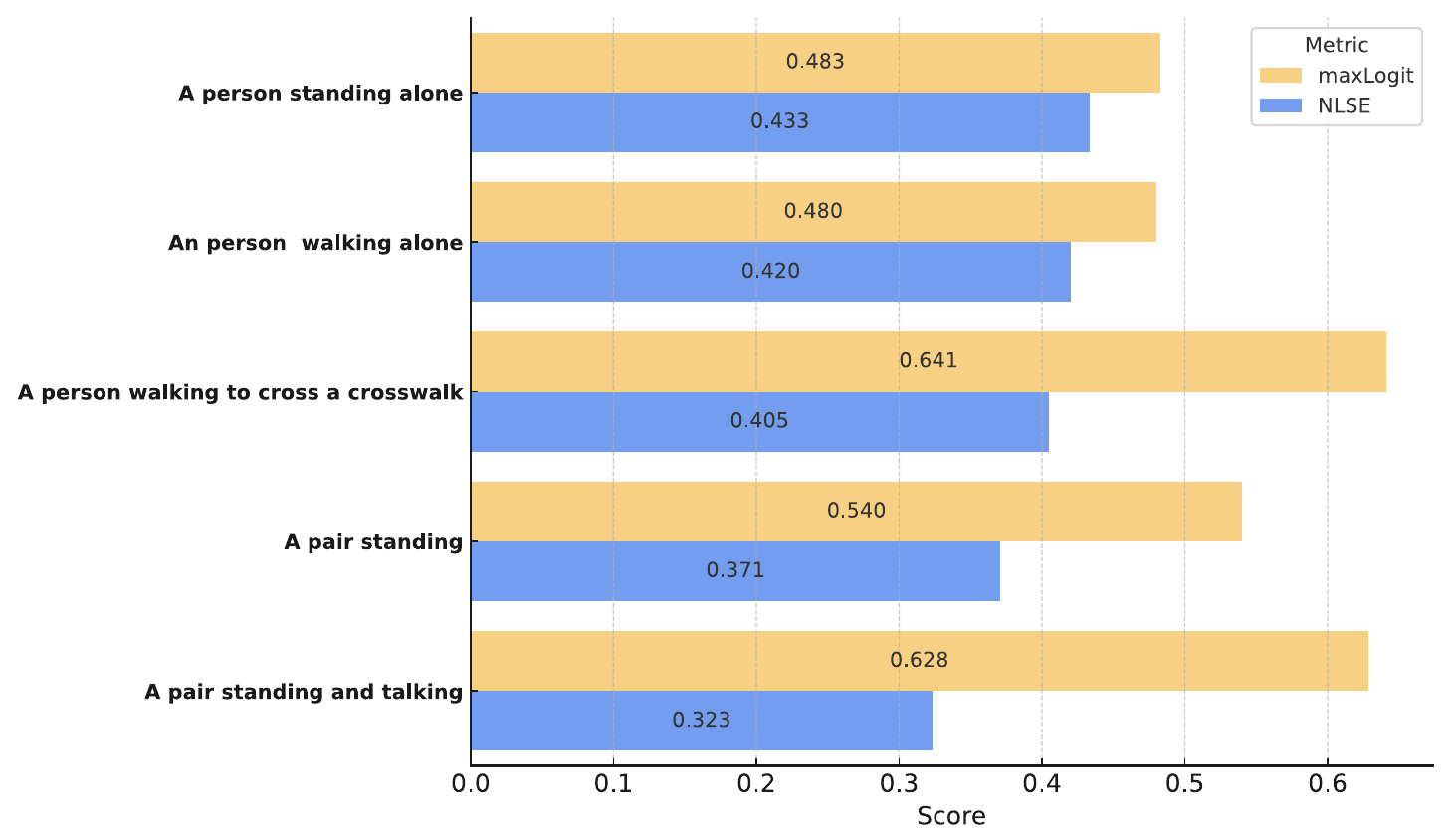}
\end{center}
   \caption{Comparison of average score of the five most frequent prompts computed using the Max-logit and N-LSE (ours). The plot shows how Max-Logit scores may be artificially inflated.}
\label{fig:NLSE}
\end{figure}

After computing the N-LSE score for all boxes, we retain only those with scores higher than 0.3 (following \cite{scalingmindrer}), resulting in 188,803 predicted boxes for SwinT Tuned model, which is roughly 0.03\% of the original set. In contrast, the Max-Logit method with the same threshold yields 2,860,823 boxes, approximately 0.4\% of the total boxes, or nearly 15 times more, underscoring the effectiveness of the N-LSE in significantly reducing the retained bounding boxes while maintaining high confidence.
We computed the average score for each prompt group (i.e., all synonymous prompts) and compared it with the average Max-Logit method over the same number of boxes. The results indicate that Max-Logit scores are often inflated, failing to reflect true model confidence in multi-label scenarios. 
The fine-tuned model Swin-T(2), adapted on COCO, demonstrated superior performance, as ``people'' is a core COCO class, providing prior knowledge of human-centric features. Since all prompts focus on human detection, this pretraining advantage improved localization accuracy.
Figure~\ref{fig:NLSE} shows comparisons for the five most frequent prompts, where Max-Logit values are often excessively high, leading to more false positives. In contrast, bounding boxes that exceed our 0.3 threshold post-N-LSE show better alignment with the ground truth.

\subsection{Analysis of DBA}
\label{subsec:localization}
Table~\ref{tab:resdba} reports a comparison between NMS-AP and DBA-AP on four OVD models. Consistently with the observations of the NMS-AP paper, our DBA approach mitigates the issue of inflated AP scores, providing a more accurate assessment of model performance by producing lower but more representative AP values. Indeed, despite the lower score, the DBA retains a number of TP instances that were suppressed by NMS (e.g. on average ~3000 occurrences for our experiments with different Grounding DINO backbones).
However, a notable trend emerges for more challenging cases, where the NMS-AP shows an AP of zero, highlighting the vulnerability of NMS-AP that was also identified in the original paper~\cite{yao2024evaluate}—where, in sub-optimal models and challenging cases, the highest confidence score does not correspond to the correct prediction. The results from both standard and DBA-AP evaluations reinforce this shortcoming, demonstrating that our DBA-AP evaluation method can recover a number of true positive predictions, leading to non-zero AP scores and a more reliable measure of the model’s performance.

To offer a more in-depth analysis of the overall performance capabilities, Table~\ref{tab:by_condition} presents the results for different conditions in which people appear in \ours: alone, in pairs, and in groups. The results exhibit the same trend as in the previous analysis, where almost all models show lower AP values with our DBA-AP, whereas the AP score collapses to zero for challenging classes, such as the class \textit{Pair}.

This comprehensive evaluation offers a clearer understanding of how these models perform under different conditions, highlighting the robustness and limitations of current approaches in detecting complex social interactions, while affirming the validity of our conclusion.

Table~\ref{tab:resranking} shows the mAP for different variants of Grounding DINO on different levels of our Benchmark when using the N-LSE or the Max-Logit approach. Due to the considerably lower number of false positive, our N-LSE approach consistently outperforms Max-Logit at almost every level.

\subsection{Performance on \ours}

Finally, we discuss the general performance of models on our benchmark task, for both NMS-AP and DBA-AP. Table~\ref{tab:resdba} shows that overall, Grounding DINO and OWL outperform the other models. Despite MDETR computing consistently higher confidence scores per box compared to Grounding DINO, its true positive detection is substantially inferior. The effectiveness of all models tends to degrade with an increase in complexity, with CSO being the hardest level to predict, showing AP values generally close to zero.

However, the overall performance of all models on our \ours{} appears significantly lower compared to other common benchmarks where those models excel. This highlights the substantial gap in detection capabilities when it comes to recognizing social activities in complex urban scenarios. 

Regarding an in-depth analysis on conditions,  Table~\ref{tab:by_condition} shows that all models struggle across all conditions, with consistently low performance. The presence of disjoint prompts in the evaluation adds further challenges for the models, as they must process and detect individuals in complex scenarios while dealing with non-overlapping categories in the prompts. These results reinforce the challenges our benchmark pose for these models and highlight the need for improved detection methods.

\begin{table}[h!]
\caption{Performance comparison of NMS-AP and DBA-AP using N-LSE confidence metrics on Alone, Pair, and Group conditions. All scores are in percentage. The numbers and symbols next to each model denote different backbones or training data: Swin-B $(b)$, Swin-T without $(t1)$ and with $(t2)$ finetuning on COCO, EfficientNet $\dagger$, and ResNet 101 $\ddagger$.}
\label{tab:by_condition}
\begin{center}
\arrayrulecolor{lightgray} 
\begin{tabular}{l l l l l }
\rowcolor{lightgray!20} 
\textbf{Model} & \textbf{Scoring} & \textbf{Alone} & \textbf{Pair} & \textbf{Group} \\ \arrayrulecolor{lightgray} \hline

\multirow{2}{*}{Gdino$^{(b)}$} 
                      & DBA-AP  & \low{2.29E-01} & \low{1.09E-01} & \low{6.09E-04} \\ \cline{2-5}
                      & NMS-AP   & {8.21E-01}  & {6.79E-01} & {6.28E-04} \\ \cline{2-5}
\multirow{2}{*}{Gdino$^{(t1)}$} 
                      & DBA-AP  & \low{1.68E-02}  & \low{4.91E-04} & \low{2.16E-07} \\ \cline{2-5}
                      & NMS-AP   & {8.14E-01}  & {5.87E-03} & {1.57E-04} \\ \cline{2-5}
\multirow{2}{*}{Gdino$^{(t2)}$} 
                      & DBA-AP  & \low{1.16E-01}  & \high{3.84E-04} & \low{1.92E-08} \\ \cline{2-5}
                      & NMS-AP   & {5.74E-01}  & {0} & {1.07E-06}  \\ \hline

\multirow{2}{*}{MDETR$^\dagger$} 
                      & DBA-AP  & \high{7.62E-04}  & \high{9.62E-05} & \low{8.70E-07} \\ \cline{2-5}
                      & NMS-AP   & {0} & {0}  & {2,43E-05} \\ \cline{2-5}
\multirow{2}{*}{MDETR$^\ddagger$} 
                      & DBA-AP  & \high{5.98E-06}  & \low{1.07E-04} & \low{1.01E-06} \\ \cline{2-5}
                      & NMS-AP   & {0}  & {6.61E-02} & {2.69E-02} \\ \hline
\multirow{2}{*}{OWL} 
                      & DBA-AP  & \low{4.96E-02} & {0} & \high{4.02E-01} \\ \cline{2-5}
                      & NMS-AP  & 4.51E-01 &  {0} & {0} \\ \cline{2-5} \hline
\multirow{2}{*}{Detic} 
                      & DBA-AP  & \high{9.78E-02} & \high{2.59E-05} & \high{2.46E-03} \\ \cline{2-5}
                      & NMS-AP  & 4.51E-02 & {0} & {0} \\ \cline{2-5} \hline
\end{tabular}
\end{center}
\end{table}

%% file: sec/06-conclusion.tex
\vspace{-10mm}
\section{Conclusion}
\label{sec:concl}
This paper introduces \ours, a novel dataset specifically curated for the detection of social activities from still images within urban environments. Employing a multi-labeling scheme, \ours comprises 934 annotated images, and more than 4,000 bounding boxes, annotated with 115 unique combinations of social activities. \ours{} comes with a new re-ranking approach, namely N-LSE, specifically designed for multi-label scenarios and OVD models, for which the effect of each token in a query is accounted in the final logit score calculations of the predicted bounding boxes. We show that N-LSE, in contrast to the Max-Logit approach in prior work, yields better performance in localization. 
With this work, we enable a more profound evaluation of OVD models in challenging real-world settings and encourage further research on applying such models in applications such as assessing human activity in still images.

%% file: sec/07-appendix.tex
\section{Appendix}
\label{sec:appen}

In this appendix we provide supplementary information about our work. 
Section \ref{subsec:label-categories} offers further details on the labels used in the study, examples of annotations are provided in Section \ref{subsec:additional-annotations}, 
while a comprehensive table listing all labels in the dataset is given in Section \ref{subsec:full-label-list}, 
additional results are presented in Section \ref{subsec:token-examples}, complemented by qualitative results In Section \ref{subsec:qualitative-res}.
In Section \ref{subsec:requirements} we provide information on the implementation details, and in Section~\ref{subsec:challenges} we discuss some of the main challenges in this work.

\textbf{Ethical Considerations:} All images used in \ours are sourced from publicly available street-view services, and our manual annotation process ensures that no personally identifiable information is disclosed. The dataset is intended solely for research purposes to advance the understanding of social interactions in public spaces.
The dataset is intended for research purposes only and should not be used for military or surveillance applications.

The dataset will be publicly available on a GitHub repository, including metadata, pre- and post-processing scripts, as well as the evaluation ecosystem. Due to dissemination restrictions, the released dataset will include metadata in CSV format, detailing fields such as \texttt{Panoid}, \texttt{latitude}, \texttt{longitude}, and \texttt{heading}. Scripts are provided to enable users to download images via respective APIs using their own API keys. The required images can be downloaded without exceeding the platforms' free-tier quotas. Future updates, including a dedicated website, are planned. Upon official release, files will be uploaded to Zenodo for long-term archival and citation (DOI to be provided).

For the submission stage, we have created an anonymous repository to share the codes and resources for this project. This ensures accessibility and transparency while maintaining anonymity for review purposes~\footnote{Paper under review. Code will be released upon acceptance}.

\subsection{Label categories}
\label{subsec:label-categories}
In the realm of social interaction recognition, the labels under the ``Activity'' category are instrumental in identifying engagement patterns and interaction types, distinguishing, for example, between conversational engagement and co-active behavior.
 
Activity labels are non-disjoint, capturing the complexity of human behavior, where multiple actions can co-occur, like \emph{talking} while \emph{pushing a stroller}.
 
We also have another category of labels, namely, ``Other'', which represents characteristics of the scene that do not fall under the previous categories and are still important for understanding the features of the urban area. For example, the label \emph{kid} can indicate a family-friendly area. 

\subsection{Annotation Strategies}
\label{subsec:additional-annotations}
\begin{figure}[h!]
\begin{center}
\includegraphics[width=0.85\linewidth]{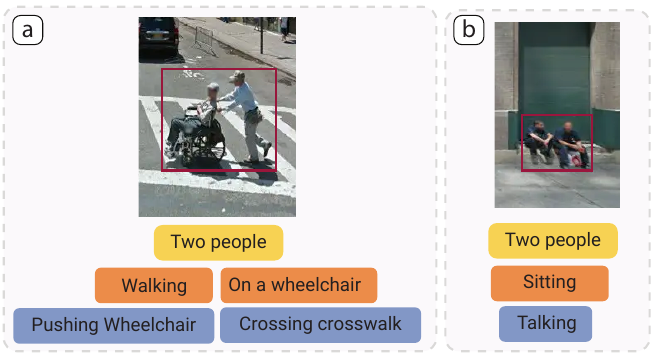}
\end{center}
   \caption{Example of rules of capture in annotation. a) Two people sitting and the stairs are not captured as an annotation. b) Two people crossing a crosswalk and one pushing a wheelchair. The wheelchair is captured in the annotation.}
\label{fig:capture}
\end{figure}

As shown in Figure~\ref{fig:capture}, for activities that are described with another non-stationary object, e.g., \emph{pushing a wheelchair} or \emph{biking}, the annotated ground truth bounding box includes the object as well as the person performing the action (see Figure~\ref{fig:capture}-a), whereas for actions without an object that is actively a part of the action, the annotated bounding box merely captures the person (see Figure~\ref{fig:capture}-b \emph{sitting}).

\subsection{Full list of labels}
\label{subsec:full-label-list}

Table~\ref{tab:list-full-labels} reports the full list of labels used during the annotation process in \ours{}. We omit some additional meta-labels, which supported the annotation process and the collection of statistics, such as ``no people'' and ``model hint''.

\begin{table*}[ht]
\centering
\begin{tabular}{|c|c|c|c|}
\hline
\textbf{Condition} & \textbf{State} & \textbf{Activity} & \textbf{Others} \\
\hline
 &  & & \\
Alone & Sitting & Dining & Pet \\
Couple & Standing & Snacking & Kid \\
Group & Walking & Talking & Police \\
& Running & Playing & Infant \\
& Biking & Shopping & Elderly \\
& On wheelchair & Hugging & Teenager \\
& Mobility aids & Taking photo & With bike \\
& Riding carriage & Talking on phone &  \\
& Riding motorcycle & Taking Taxi & \\
& & Pet interactions & \\
& & Street vendors & \\
& & Phone interaction & \\
& & Waving to camera & \\
& & Pushing stroller & \\
& & Sport activities & \\
& & Crossing crosswalk & \\
& & Pushing wheelchair & \\
& & Working with laptop & \\
& & Construction workers & \\
& & Pushing shopping cart & \\
& & Waiting in bus station & \\
& & At petrol/gas station & \\
& & Public service/cleaning & \\
& & Load/unload packages from car/truck & \\
&&&\\
\hline
\end{tabular}
\caption{Full list of labels in \ours{} divided by category}
\label{tab:list-full-labels}
\end{table*}

\subsection{Sanity Rules for Annotation Cleaning}
\label{subsec:sanity-rules}
To make sure that all the annotated labels for bounding boxes are correct, we performed a sanity-check using a predefined set of sanity rules. In the following, we summarize the full set of rules we considered at this stage:

\begin{enumerate}
    \item Each bounding box must have a condition label, unless it is a ``pet'';
    \item Each bounding box must have at least one state label, unless it is a ``pet'';
    \item Each bounding box can only have one condition label associated, e.g., ``alone'' and ``group'' cannot appear together;
    \item If a bounding box is associated with the ``alone'' condition, then it can only have one state label associated, e.g., ``alone walking running'' is not allowed;
     \item If a bounding box is associated with the ``couple/two person'' condition, then it can only have two state labels associated, e.g., ``couple walking sitting running'' is not allowed;
    \item If a bounding box is associated with the ``shopping'' activity, then state should include either one of ``sitting'' or ``standing'' labels;
    \item If a bounding box is associated with the ``street vendors'' activity, then state should include either one of ``sitting'' or ``standing'' labels;
    \item If a bounding box is associated with the ``load/unload packages'' activity, then state should include either one of ``sitting'' or ``standing'' labels;
    \item If a bounding box is associated with the ``waiting in bus station'' activity, then state should include either one of ``sitting'' or ``standing'' labels.
\end{enumerate}

\subsection{Additional Results}
\label{subsec:token-examples}
\begin{figure*}[t]
    \centering
    \includegraphics[width=\linewidth]{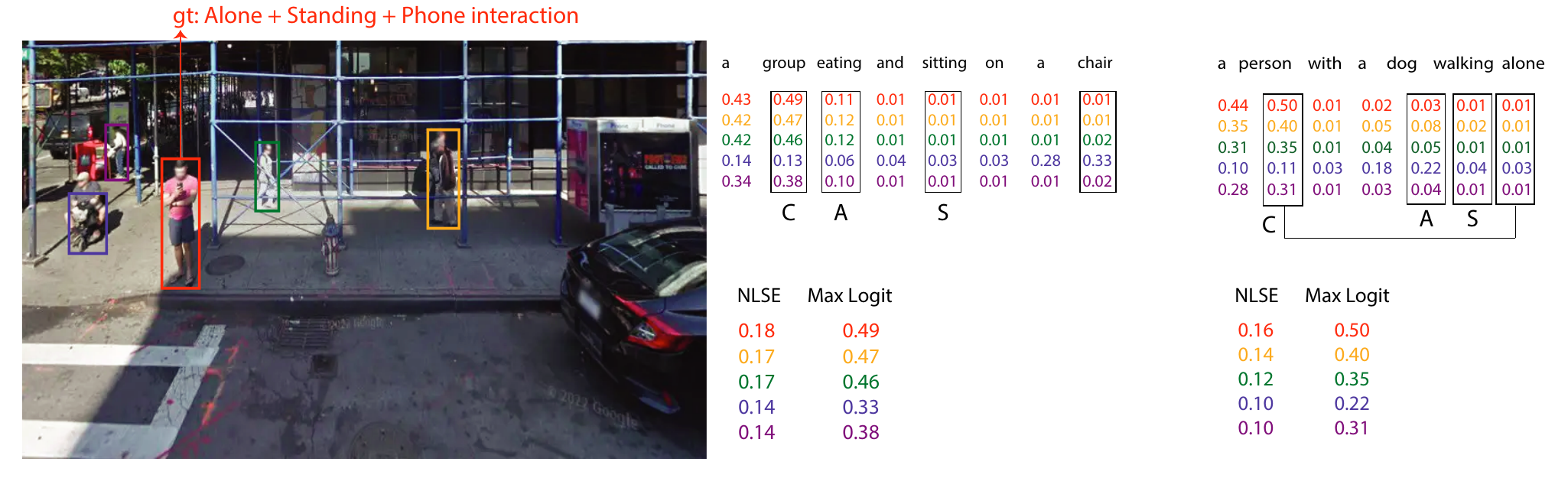}
    \caption{An example of the top five predictions of the Grounding DINO model for two distinct prompts on the same image is provided. Top tables present the model's confidence in the presence of the tokens within each box. The selected tokens used to compute the N-LSE metric are highlighted with boxes annotated by C:condition, S:state, A: activity. Bottom tables display the overall score for each color-coded box, comparing N-LSE on selected tokens (ours), and the maximum logit of all tokens. }
    \label{fig:token}
\end{figure*}

\textbf{Selecting Relevant Logits}. Grounding DINO uses the BERT model for tokenization. We keep the mapping between logits and tokens and their category of condition, state, activity. Using this mapping, we only keep the relevant tokens in our metric calculation. Figure~\ref{fig:token} shows our metric being applied to relevant tokens, as well as the Max-logit.
fIn both prompts, one target (the red box) was predicted with the highest confidence. The ground truth for that target comprises the following labels: \emph{C: Alone + S: Standing + A: Phone interaction}. In this example, we showcase how the same target is assigned two disjoint conditions, with high confidence. The same individual is returned as the highest confidence prediction for first prompt: ``a group eating and sitting on a chair'', with 49\% confidence in representing a ``group'', and 11\% eating. While in the second prompt has a matching condition only, ``alone'', which was returned by the model with 50\% confidence. All predictions have a fairly close confidence in the target representing disjoint conditions, highlighting the low understanding of the model in interpreting the condition in this image. 

None of the people in this image match any of our queries. However, using the max log score, for the first prompt (Figure~\ref{fig:token}-top), all five boxes would pass the 0.3 threshold and be counted as likely candidates. However, using our score (N-LSE), none of the boxes would be selected. The same applies for the other prompt. There is a notable difference between the two scores, highlighting the important role of taking relevant query terms into account.

\subsection{DBA}

Key features of \algname include:

\begin{itemize} \item \textbf{Confidence Range Consideration}: Instead of selecting only the box with the highest confidence, \algname retains all boxes whose confidence scores are within a specified threshold (\texttt{score\_thr}) of the maximum confidence in the group. This approach prevents the unnecessary suppression of true positives that have slightly lower confidence scores. \item \textbf{Semantic Consistency Check}: \algname examines the predicted labels for disjointness in critical categories (e.g., condition or state). If overlapping boxes predict contradictory labels (e.g., \emph{sitting} vs. \emph{walking}), they are flagged as false positives. \end{itemize}

\subsection{Qualitative Results}
\label{subsec:qualitative-res}
As a prompt increases in level from \emph{condition} to \emph{condition, state, activity, and others,} the likelihood that the prompt contains labels that the model has a low confidence on from its training increases, lowering the computed score for the box. The outcome is that the most basic-level prompts are overrepresented among the predictions that pass score-based filters, and high-level prompts are extremely uncommon. \emph{Condition} prompts accounted for less than 2\% of the total prompts generated, but were 20\% of the bounding boxes that passed the initial thresholding on score. Conversely, when more conventionally
determining the score by the maximum logit for the box, higher-level prompts have more logits and therefore always result in higher representation in the predictions that pass the threshold. 

When a prompt includes an object that is among the pre-trained vocabulary, the model can more easily detect and localize it. This is a case where contextual cueing leads to better predictions. For instance, when we query for ``group of people sitting'', the model less frequently finds the correct target, but the prompt ``groups of people sitting on a chair'' can lead to a better prediction.

The most challenging part for the models was recognizing \emph{state}. The confidence of the model in associating the area inside each box with the labels in \emph{state} group is very low across all images and all set of queries.

To further analyze the model's understanding of people's states (sitting, standing, walking, etc.) we prompt it using its native Max-logit scoring and the 0.3 threshold.  

Here, we used variations of our original prompt ``a group of people sitting on a bench'' : ``a group of people standing on a bench''; and ``a group of people running on a bench''. These prompts do not have semantically valid \emph{state verbs} and are not among our set of prompt list.

In all three cases, one target was in common and had the highest confidence, as shown in Figure~\ref{fig:qual}. When prompted \emph{people sitting on a bench}, the model returned one result  44\% confidence, however, the model assigned higher confidence to the same target with \emph{people standing on a bench} with 52.98\% confidence
and 53.08\% confidence in the box showing \emph{people running on a bench}. The Max-logit method results in false positive predictions with very high confidence and undermine the actual context of the query by allowing the logit with the maximum confidence to represent the whole query. 

Our findings also highlight the need for the incorporation of uncertainty estimation techniques during model fine-tuning and training to mitigate the risk of overconfident false predictions.

\subsection{Implementation Details}
\label{subsec:requirements}
The generation of all the predictions with Grounding DINO, MDETR, Detic and OWL takes around twelve, eight, twenty and six hours respectively on one H100 with 80GB of memory. The generation of the results on an Intel(R) Xeon(R) Platinum 8480CL takes around ten minutes for each model. 

We used the Open Grounding DINO implementation \footnote{\url{https://github.com/longzw1997/Open-GroundingDino}}. Our inference was conducted using the  configuration from the official repository with Swin-T backbone, pre-trained on O365, GoldG, and Cap4M datasets. For MDETR \footnote{\url{https://github.com/ashkamath/mdetr}} and Detic \footnote{\url{https://github.com/facebookresearch/Detic}} we used the official repository and checkpoints, whereas for OWL we employed the ViT Patch 32 version from the Hugging Face hub\footnote{\url{https://huggingface.co/docs/transformers/model_doc/owlvit}}.

\subsection{Notes on challenges}\label{subsec:challenges}
Existing OVDs exhibit a number of challenges. They often struggle with semantic consistency across diverse inputs, showing limited adaptability to novel or unseen categories, and can suffer from high computational costs during inference. Additionally, these models may demonstrate sensitivity to slight variations in input phrasing, leading to inconsistent performance. The calibration of their predictive confidence, especially in out-of-distribution scenarios, remains suboptimal, frequently resulting in overconfident predictions that do not accurately reflect their actual accuracy.

Aside from the challenging nature of human activity and interaction detection, the lower quality of large-scale publicly available street-level images impact the detection results. On top of that, the anonymization process to blur faces creates artifacts that may affect other people in the scene, making them more difficult to be detected. 